\documentclass[NewProceedings,letterpaper]{ascelike-new}
\usepackage{booktabs}
\usepackage[utf8]{inputenc}
\usepackage[T1]{fontenc}
\usepackage{lmodern}
\usepackage{graphicx}
\usepackage[figurename=Fig.,labelfont=bf,labelsep=period]{caption}
\usepackage{subcaption}
\usepackage{amsmath}
\usepackage{newtxtext,newtxmath}
\usepackage[colorlinks=true,citecolor=black,linkcolor=black]{hyperref}

\usepackage[usenames,dvipsnames,svgnames,table]{xcolor}
\usepackage{algorithm, algorithmicx, algpseudocode}

\usepackage{svg}
\usepackage[none]{hyphenat}
\NameTag{\textcolor{Grey}{Xu. 12/2024}}
\begin{document}

\title{\textit{Seeing with Partial Certainty}: Conformal Prediction for Robotic Scene Recognition in Built Environments}

\author[1]{\textbf{Yifan Xu, S.M. ASCE}}
\author[2]{\textbf{Vineet Kamat, F. ASCE}}
\author[3]{\textbf{Carol Menassa, F. ASCE}}

\affil[1]{Department of Civil and Environmental Engineering, University of Michigan, Ann Arbor, MI, 48109-2125. email: yfx@umich.edu}
\affil[2]{Department of Civil and Environmental Engineering, University of Michigan, Ann Arbor, MI, 48109-2125. email: vkamat@umich.edu}
\affil[3]{Department of Civil and Environmental Engineering, University of Michigan, Ann Arbor, MI, 48109-2125. email: menassa@umich.edu}

\maketitle

\begin{abstract}
\hfill 
\par 

In assistive robotics serving people with disabilities (PWD), accurate place recognition in built environments is crucial to ensure that robots navigate and interact safely within diverse indoor spaces. Language interfaces, particularly those powered by Large Language Models (LLM) and Vision Language Models (VLM) hold significant promise in this context, as they can interpret visual scenes and correlate them with semantic information. However, such interfaces are also known for their \textit{hallucinated} predictions. In addition, language instructions provided by humans can also be ambiguous and lack precise details about specific locations, objects, or actions, exacerbating the hallucination issue. In this work, we introduce \textit{Seeing with Partial Certainty} (SwPC) — a framework designed to measure and align uncertainty in VLM-based place recognition, enabling the model to recognize when it lacks confidence and seek assistance when necessary. This framework is built on the theory of \textit{conformal prediction} to provide statistical guarantees on place recognition while minimizing requests for human help in complex indoor environment settings. Through experiments on the widely used richly-annotated scene dataset Matterport3D, we show that SwPC significantly increases the success rate and decreases the amount of human intervention required relative to the prior art. SwPC can be utilized with any VLMs directly without requiring model fine-tuning, offering a promising, lightweight approach to uncertainty modeling that complements and scales alongside the expanding capabilities of foundational models.

\end{abstract}

\section{Background and Challenge}
\label{sec:intro}

Assistive robots have the potential to play a crucial role for people with disabilities (PWD) by providing essential support that enhances independence, mobility, and quality of life ~\cite{Quesada2022}. Place recognition (e.g., kitchen, living room) is a critical capability for assistive robots designed to support PWD, as it enables robots to understand and navigate the spaces in which they operate~\cite{Karasfi2011}. For individuals with mobility or cognitive impairments, reliable place recognition allows assistive robots to accurately interpret environmental cues, understand spatial layouts, and follow commands related to specific locations or tasks. This is particularly essential in environments like homes~\cite{sheng2024nycindoorvprlongtermindoorvisual} or care facilities~\cite{Zeng2018}, where users may need assistance moving between rooms or locating objects, and where a robot’s awareness of its surroundings directly impacts its ability to provide safe and effective support. 

Recently, language has been proven to be an effective link between intelligent systems and humans and can enable robot autonomy in complex human-centered environments~\cite {ahn2022icanisay}. The recent success of Visual Language Models (VLM) demonstrates that open-vocabulary capabilities are crucial for place recognition~\cite{woo2024}.  In traditional close-vocabulary settings, recognition is limited to a fixed set of labeled categories, restricting the robot’s ability to identify new or dynamically changing locations. In contrast, VLMs can interpret new terms and associate them with visual characteristics, enabling the robot to adapt to new locations or objects based on natural language descriptors~\cite{Brose2010,Guadarrama2015,Zang2022}. 

However, VLMs have a significant challenge, which is their tendency to \textit{hallucinate} — producing outputs that, while seemingly plausible, are confidently incorrect and disconnected from reality~\cite{Farquhar2024}. Such hallucination causes the model to misidentify locations with high confidence, which can lead to errors in navigation and contextual understanding critical for assistive applications ~\cite{Jha2023,zhang2024}. Furthermore, natural language instructions in real-world environments often carry a high degree of inherent or unintentional ambiguity from humans, increasing the likelihood of hallucinations occurring~\cite{Hatori2018}.  For example, a robot tasked with guiding people to their bedrooms may be asked to \textit{"go to the bedroom"}; if there are multiple bedrooms in the whole indoor environment, the instructions are ambiguous. Instead of acting in an ambiguous situation and potentially going to the wrong room, the robot should recognize its uncertainty and seek clarification (e.g., request additional information about the room to ensure accuracy and reduce frustration among users).

How can we let the intelligent vision language systems \textit{know when they are uncertain}? Accurately modeling and accounting for uncertainty is a longstanding challenge for robots that operate reliably in unstructured and novel environments. Most prior work overlooks the uncertainty in VLMs~\cite{ahn2022icanisay} and instead focuses on extensive prompt engineering to reduce hallucinations~\cite{huang2022}, a process that demands meticulous crafting of prompts to prevent the robot from becoming overly reliant on requesting assistance. KnowNo ~\cite{ren2023robotsaskhelpuncertainty}  recently framed the problem of deciding when a robot should seek assistance as one of uncertainty alignment regarding the robot task planning. However, these works have not explored uncertainty alignment in open-vocabulary room or scene classification.

\section{Statement of Contribution}
\label{sec:Contribution}
\par 

In this work, we propose a new uncertainty alignment framework SwPC - \textit{Seeing with Partial Certainty }- a framework that aligns the place recognition uncertainty with Conformal Prediction (CP) ~\cite{ACP}.  Our key finding is that by utilizing CP at the same prediction success rate, our proposed SwPC will require less human help,
which will improve the efficiency of place recognition. Our main contributions are the following: 
\begin{itemize}
\item We introduce an uncertainty alignment framework for VLM place recognition called SwPC utilizing CP to align uncertainty and confidence.
\item We provide a calibration method based on CP that can be used for uncertainty alignment in any open vocabulary detection task.
\item We evaluate our proposed CP on a widely used scene dataset, Matterport3D, and compare the result with previous uncertainty alignment methods. We show that our proposed CP approach increases the success rate and reduces the amount of help needed as compared to our baselines.
\end{itemize}

\section{Preliminary Knowledge}
\label{sec:prelim}

\begin{figure*}[t]
    \centering
    \includegraphics[width =1.0 \textwidth]{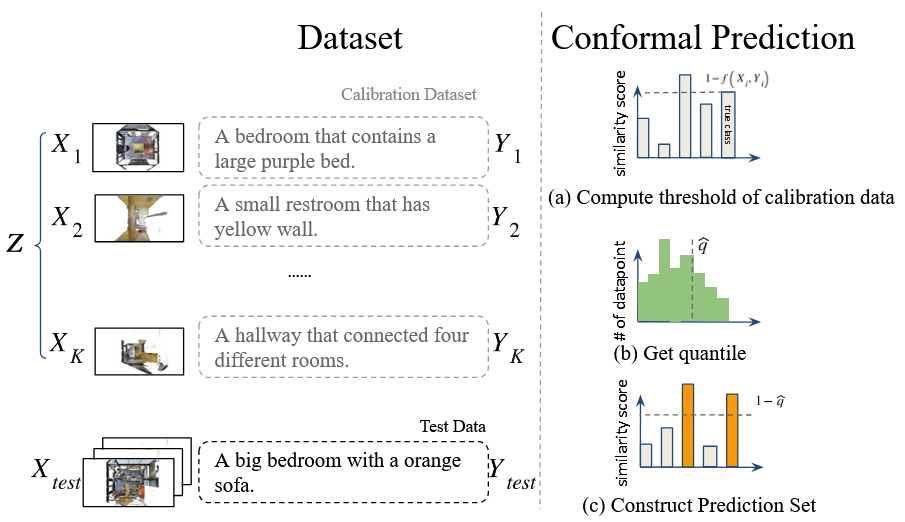}
    \caption{Illustration of conformal prediction (CP).} 
    \label{fig:CP pipeline}
\end{figure*}

Conformal Prediction (CP) is a straightforward way to generate prediction sets for any model. As illustrated in the right plot of Fig.~\ref{fig:CP pipeline}, suppose we have a set of $K$ different place descriptions from user $Y_i\in \mathcal{Y}$ and want to use it to match $K$ different rooms in a large indoor region from a set of top-down view image $X_i \in \mathcal{X}$ using VLM classifier $f$ that outputs estimated probabilities (softmax scores) for each class. Then we reserve a moderate number of pairs of room images and description $(X_1, Y_1), ..., (X_n, Y_n)$ from the whole dataset for use as calibration data, which is denoted as $\mathcal{Z}=\{Z_i=(X_i, Y_i)\}_{i=1}^n$. Using the calibration data and VLM classifier $f$, we seek to construct a \textit{prediction set} of possible description $\mathcal{C}(X_{test}) \subset \{1, ..., K\}$ that satisfy the following 

\begin{equation}
    \label{eq: CP}
    \begin{aligned}
    1-\alpha \leq \mathbb{P}(Y_{test} \in \mathcal{C}(X_{test})) \leq 1-\alpha + \frac{1}{n+1}
    \end{aligned}
\end{equation}

where $(X_{test}, Y_{test})$ is a fresh test point from the same distribution, and $\alpha \in [0, 1]$ is a user-chosen error rate. The probability that the prediction set contained the true description is almost the same as $1-\alpha$. We will first need to calibrate the dataset to get the prediction dataset. We construct the set of \textit{non-conformity} score to be $\mathcal{S}=\{ s_i =  1 - f(X_i, Y_i) \}_{i=1}^{n}$. Then CP perform calibration by defining $\hat{q}$ as $\frac{\lceil (n+1)(1-\alpha)\rceil }{n}$ percentage quartile of $\mathcal{S}$. Lastly, we will construct the prediction set $\mathcal{C}(X_{test})=\{Y_{test} \in \mathcal{Y} |f(X_{test}, Y_{test}) \geq 1-\hat{q}\}$. According to Theorem 1, the generated prediction set ensures that the coverage guarantee in Eq.~\eqref{eq: CP} holds.

\subsection{Theorem 1}
\label{theorem1}
(Conformal calibration coverage guarantee). Suppose $(X_i, Y_i), i=1,...,n$ and $(X_{test}, Y_{test})$ are independently and identically distributed. Then define $\hat{q}$ as 

\begin{equation}
    \begin{aligned}
    \hat{q}=\inf\{q: \frac{|\{i:1 - f(X_i, Y_i)\} \leq q|}{n} \geq \frac{\lceil (n+1)(1-\alpha)\rceil }{n}\}
    \end{aligned}
\end{equation}

and the resulting prediction set as 

\begin{equation}
    \begin{aligned}
    \mathcal{C}(X) = \{Y : 1 - f(X, Y) \leq \hat{q}\}
    \end{aligned}
\end{equation}

Then,

\begin{equation}
    \begin{aligned}
    P(Y_{test} \in \mathcal{C}(X_{test})) \geq 1-\alpha
    \end{aligned}
\end{equation}

\section{methodology}
\label{sec:methodology}

\begin{figure*}[t]
    \centering
    \includegraphics[width = 1.1\textwidth]{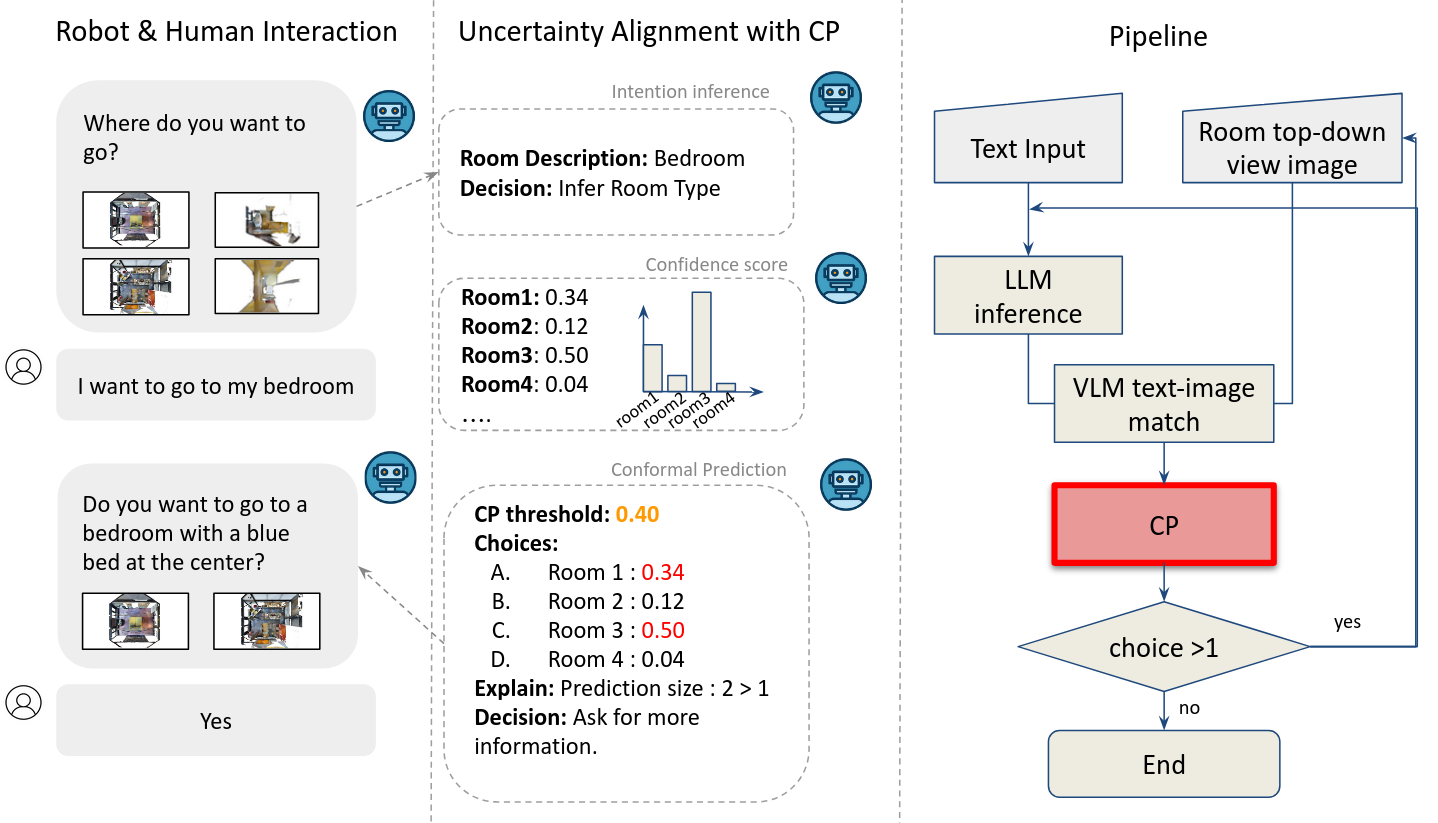}
    \caption{Illustration of SwPC pipeline.}
    \label{fig:SwC pipeline}
\end{figure*}

In this work, we consider the following problem statement: \textit{Given a set of different rooms, which is represented by different top-down views, and a user-provided instruction, how to recognize the intended room accurately while resolving any ambiguities as needed through communicating with the user and minimizing the amount of dialogue.} As illustrated in Fig.~\ref{fig:SwC pipeline}, the process begins with the LLM inference module, where we use prompt engineering to extract a room description from user-provided input. Next, the VLM computes cosine similarity scores between the user’s room description and various top-down room images. Based on a predefined error rate, CP is then applied to generate a prediction set of possible rooms. If the prediction set contains more than one candidate, additional user input is required to clarify the intended destination.

\subsection{Calibration Dataset Collection}

We will first need to get the calibration dataset to calibrate the uncertainty. The steps of construction of the calibration dataset are below:
\begin{enumerate}
  \item We utilize a well-trained Multimodal Large Language Model (LMM) to generate descriptions $X_i, i=1,...n$ of different rooms from top-down images $Y_i, i=1,...,n$.
  \item We split the image-text pair data into two parts, one part for fine-tuning VLM and another part for CP calibration. 
  \item After fine-tuning VLM, we utilize the fine-tuned VLM to get the cosine similarity score between different descriptions and images. 
  \item We sort the cosine similarity score and form the non-conformity score - image - description pairs as the raw datasets used for CP calibration. Inspired by the Oracle algorithm~\cite{gibbs2024}, the non-conformity score is described as the similarity score. We define $\pi(X_i)=\{\pi_1(X_i), ... ,\pi_n(X_i)\}$ to be the permutation of $\{1,...,n\}$ that sorts $\{f(X_i, Y_1),...,f(X_i, Y_n)\}$ from most likely to least likely. The dataset is described below: 
  \begin{equation}
    \begin{aligned}
    \mathcal{D}=\{(s(X_i,Y_j),X_i,Y_j ) | s(X_i,Y_j) = 1 - f(X_i, Y_{\pi_i(X_i)}), i=1...,n, j=\pi_1(X_i), ... ,\pi_n(X_i)\}
    \end{aligned}
  \end{equation}
  \item We include classes with true labels. According to the theory of conformal prediction, we form the calibration dataset as below:
  \begin{equation}
    \begin{aligned}
    \hat{\mathcal{D}}=\{(s(X_i,Y_j),X_i,Y_j) | (s(X_i,Y_j),X_i,Y_j )\in \mathcal{D}, i=1,...,n, j=\pi_i(X_i)\}
    \end{aligned}
  \end{equation}

\end{enumerate}

\subsection{Conformal Prediction}

After we get the calibration dataset, we will perform conformal prediction to get our prediction set. Similar as the step 4 of calibration dataset collection, we define $\pi(X_{test})=\{\pi_1(X_{test}), ...,\pi_n(X_{test})\}$ to be the permutation of $\{1,...,n\}$ that sorts $\{f(X_{test}, Y_1),...,f(X_{test}, Y_n)\}$ from most likely to least likely. Then, we define a score function as below:

\begin{equation}
\begin{aligned}
s(X_{test}, Y_{\pi_{k}(X_{test})}) = 1 - f(X_{test}, Y_{\pi_{k}(X_{test})}), \text{where}~k=1,...,n
\end{aligned}
\end{equation}

The next step is to set the quartile value $\hat{q}$ of the calibration dataset, which is the same as in any conformal prediction.

\begin{equation}
\begin{aligned}
\hat{q}=\text{Quartile}(\hat{\mathcal{D}}, \frac{\lceil (n+1)(1-\alpha)\rceil }{n})
\end{aligned}
\end{equation}

where $\hat{D}$ is the calibration dataset that is from the previous calibration dataset generation step. $\alpha \in [0,1]$ is the people defined error rate. Having done so, we will form the prediction set $\mathcal{C}(X_{test})$ as below:

\begin{equation}
\begin{aligned}
\mathcal{C}(X_{test}) &=\{Y_{\pi_1(X_{test})},...,Y_{\pi_k(X_{test})} \}, \\
\text{where}~k&=\sup\{k^{'}:s(X_{test}, Y_{\pi_{k^{'}}(X_{test})}) \leq \hat{q}\} + 1
\end{aligned}
\end{equation} 
\section{Experiment and Results}
\label{sec:results}
\par 


We evaluate our proposed SwPC framework in a diverse set of indoor built environments. In the experiment below, we have demonstrated the effectiveness of reaching the success rate while minimizing the rate of asking humans for help. The dataset, baseline, and experiment results are introduced below. We utilize LLaVA~\cite{liu2023llava} as the multimodal LLM to generate the description of the rooms and use LongCLIP~\cite{zhang2024longclip} as the VLM backbone to do text-image matching. 

\subsection{Dataset}

We conducted our experiment on the Matterport3D dataset~\cite{Matterport3D}, where all multi-floor scenes were segmented into single-floor scenes, resulting in 189 scenes. Excluding those used for fine-tuning the VLM and calibration of CP, we selected 43 scenes, which include 1504 rooms, to serve as the test set for room segmentation. Since we are generating descriptions for these rooms and using the same VLM model to generate the similarity score, the test data and calibration data follow the same distribution. 

\subsection{Baseline}

Our baselines for comparison are Prompt Set, No Help Set, and Binary Set. The introduction of these baselines is shown below:

\begin{itemize}
    \item \textbf{Prompt Set}: Prompts the multimodal LLM to output the prediction set directly. (e.g. “Prediction set: [B, D]”) 
    \item \textbf{Binary Set}: Prompts the multimodal LLM to directly output a binary indicator of uncertainty (e.g., “Certain/Uncertain: Certain”) which is used in other LLM-based planning work~\cite{huang2022} for triggering human intervention.
    \item \textbf{No Help Set}: Always uses the highest similarity score directly from the VLM without creating a prediction set or asking for human intervention.
\end{itemize}

\subsection{Metrics}
\begin{figure}[!tbp]
  \begin{subfigure}[b]{0.55\textwidth}
    \includegraphics[width=\textwidth]{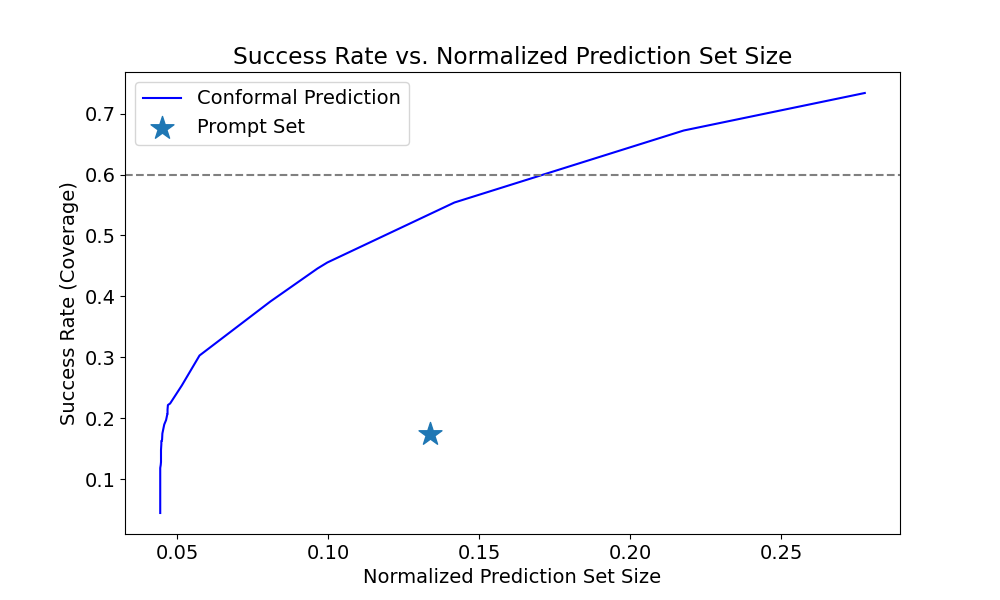}
    \caption{Success Rate vs. Normalized Prediction Set Size}
    \label{fig:f1}
  \end{subfigure}
  \hfill
  \begin{subfigure}[b]{0.55\textwidth}
    \includegraphics[width=\textwidth]{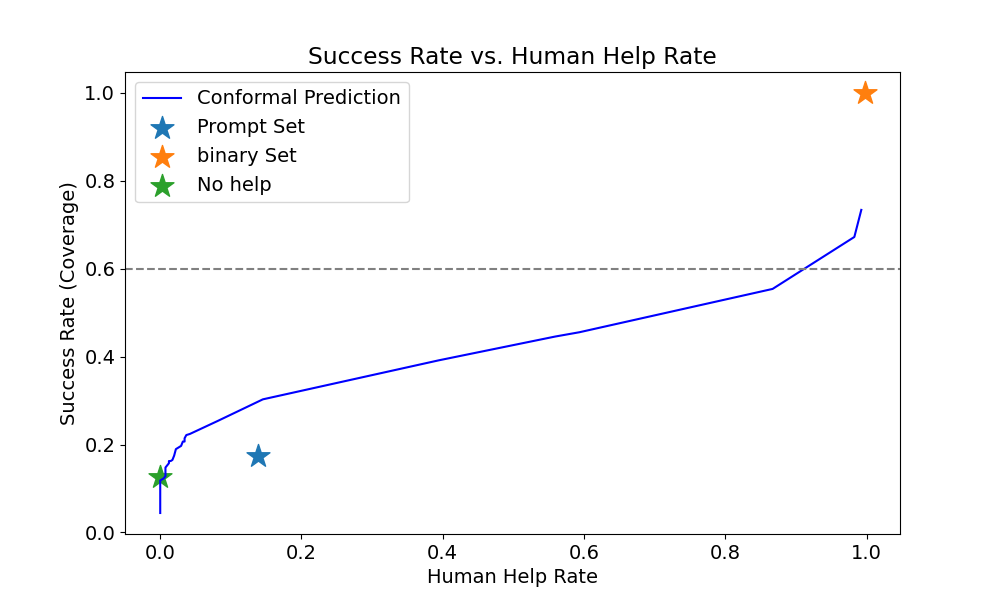}
    \caption{Success Rate vs. Human Help Rate}
    \label{fig:f2}
  \end{subfigure}
  \caption{Comparison of task success rate vs average prediction set size (Left) and vs. human help rate (Right) of Matterport3D dataset averaged over the three settings. 1504 rooms are evaluated for each method. $\alpha$ is varied from 0 to 1 for CP. Binary and No Help are not shown on the left since prediction sets are not provided.}
  \label{fig:plot1}
\end{figure}

\begin{figure*}[t]
    \centering
    \includegraphics[width = \textwidth]{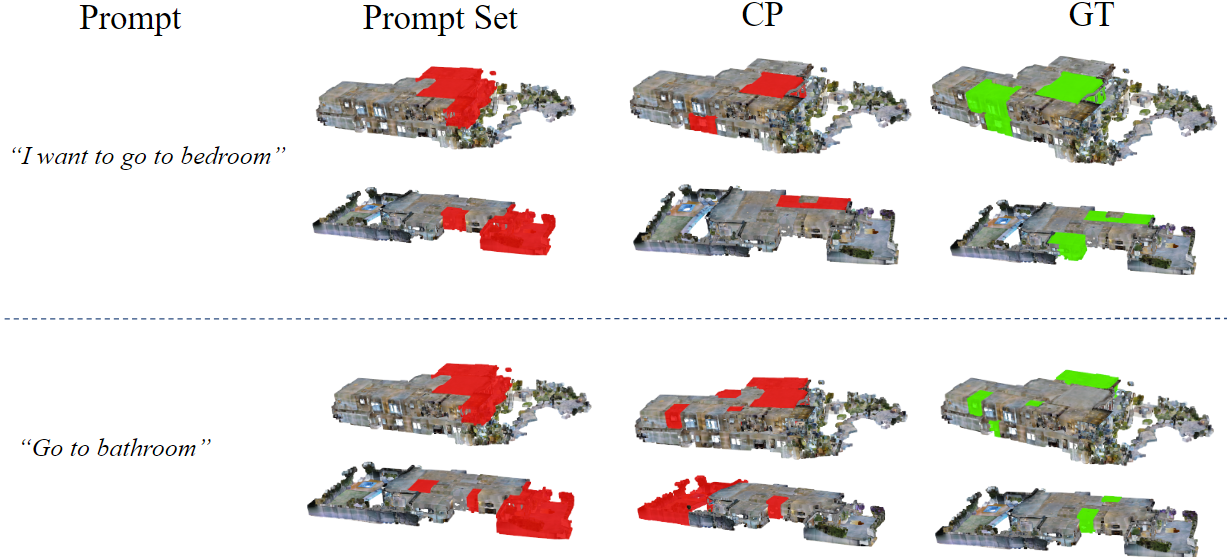}
    \caption{Qualitative comparison between Prompt Set and CP Set over Matterport3D. From the qualitative result, the overlapping area between ground truth and CP prediction is greater than the baseline Prompt Set. Binary and No Help Set are not shown on the left since prediction sets are not provided. }
    \label{fig:qualitative result}
\end{figure*}
We evaluate efficiency by comparing the success rate against the prediction set size and help rate between our proposed SwPC and baseline methods. A method is considered more efficient if, at the same success rate, it achieves a lower help rate or a smaller prediction set size. For a given room description, the human help rate is one if the prediction set size is more than 1. Since \textbf{No Help Set} always uses the highest score, the prediction size is always 1, and the human help rate is always 0. For \textbf{Prompt Set} and \textbf{Binary Set}, it yields a single success rate, a singular help rate, and an average prediction set size. For CP evaluation, we set different error rates $\alpha$ from 0 to 1 and calculate the success rate, prediction set size, and human help rate, respectively. Since the total number of rooms varies for different scenes, the normalized prediction set size is calculated instead. 

\subsection{Experiment Result}

In Fig.~\ref{fig:plot1}, we set up different error rates $\alpha$ and show the curves of task success rate vs. prediction set size and human help rate averaged over CP settings.

The results from the plots highlight the advantages of the proposed CP method over the Prompt Set, Binary Set, and No Help Set approaches. Fig.~\ref{fig:f1} shows that At the Prompt Set’s Normalized Prediction Set Size (0.13), the success rate of CP (53\%) exceed that of Prompt Set (18\%). CP effectively balances prediction set size and success rate (coverage), achieving higher success rates by dynamically adjusting prediction set sizes based on calibrated confidence levels. Since Binary Set directly generates certain/uncertain as an indicator for triggering human's help and No Help Set is always executed without human intervention, they do not provide prediction set. 

In contrast, the Prompt Set suffers from issues like VLM hallucination and bias, resulting in low and inflexible success rates, while the Binary Set performs well at specific success levels but lacks flexibility and cannot generate prediction sets for user feedback. Fig.~\ref{fig:f2} demonstrates that CP also optimally balances human help rates and success rates, achieving superior success-to-help ratios across varying error thresholds. From the result, CP provides a flexible, continuous trade-off curve between human help and success rate: the No Help set achieves roughly 12\% success with zero human intervention, Prompt Set needs 15\% human help rate and yields about 19\% success, Binary Set have perfect success rate with 100\% human help rate. Unlike the Prompt Set, which offers only one point of trade-off, CP can be optimized so that an equivalent level of human assistance yields significantly better results (31\%). Similarly, compared to the extremes of No Help (low success with no help) or Binary Set (perfect success, but full help with high cost), CP enables achieving high success rates at a fraction of the human cost. 
Overall, CP proves to be the most adaptable and effective approach for ensuring high success rates with minimal human intervention compared with other baseline methods.

In addition, we performed a qualitative comparison between the Prompt Set and the CP Set, as shown in Fig.~\ref{fig:qualitative result}. The evaluation focused on two common scenarios where ambiguous instructions require multiple rooms to be included in the prediction set. The Prompt Set relies on the LLM to directly generate the prediction set. However, due to the LLM's tendency to hallucinate and exhibit biases—such as prioritizing the first few choices rather than selecting the most likely ones—the accuracy of the Prompt Set is lower compared to the CP Set.


\section{Conclusion}
\label{sec:conclusion}
\par In conclusion, our study demonstrates the effectiveness of the Seeing with Partial Certainty (SwPC) framework in addressing challenges in place recognition for assistive robotics in indoor environments. By leveraging conformal prediction theory, SwPC provides reliable statistical guarantees on model predictions and mitigates the issue of hallucinations in VLM. It enables uncertainty-aware decision-making, allowing the model to identify situations where additional human assistance is needed, thus improving both success rates and efficiency. Importantly, SwPC operates without requiring fine-tuning of underlying VLMs, making it a scalable and lightweight solution for uncertainty modeling. These results highlight the potential of SwPC to enhance the safety and adaptability of assistive robots, marking a significant step forward in deploying language-based interfaces for real-world applications.
\clearpage
\section{Acknowledgements}
\label{sec:acknowledgement}

The work presented in this paper was supported financially by the United States National Science Foundation (NSF) via grant SCC-IRG 2124857. The support of the NSF is gratefully acknowledged.


%
%
\bibliography{ascexmpl-new}

\end{document}